\documentclass[10pt,twocolumn,letterpaper]{article}

\usepackage[pagenumbers]{cvpr} % To force page numbers, e.g. for an arXiv version

\usepackage[accsupp]{axessibility}

\usepackage[dvipsnames]{xcolor}

\usepackage{multibib}

\definecolor{cvprblue}{rgb}{0.21,0.49,0.74}
\usepackage[pagebackref,breaklinks,colorlinks,citecolor=cvprblue]{hyperref}
\usepackage{xcolor}
\usepackage{caption}
\usepackage{lipsum}
\usepackage{enumerate}
\usepackage{subcaption}
\usepackage{blindtext}

\usepackage{algorithm}
\usepackage{algorithmicx}
\usepackage{algpseudocode}

\def\paperID{12900} % *** Enter the Paper ID here
\def\confName{CVPR}
\def\confYear{2024}

\title{Feature 3DGS: Supercharging 3D Gaussian Splatting to Enable Distilled Feature Fields}

\author{Shijie Zhou$^1$ \quad  Haoran Chang$^1$\thanks{Equal contribution.} \quad Sicheng Jiang$^1$\footnotemark[1] \quad Zhiwen Fan$^2$ \quad  
Zehao Zhu$^2$ \quad Dejia Xu$^2$ \\ \quad Pradyumna Chari$^1$ \quad Suya You$^3$ \quad Zhangyang Wang$^2$ \quad Achuta Kadambi$^1$\\
\normalsize{$^1$University of California, Los Angeles \quad
$^2$University of Texas at Austin \quad
$^3$DEVCOM Army Research Laboratory}}

\begin{document}
%\maketitle

\twocolumn[{%
\renewcommand\twocolumn[1][]{#1}%
\maketitle
    \captionsetup{type=figure}
    %\vspace{-9mm}
    \includegraphics[width=\textwidth]{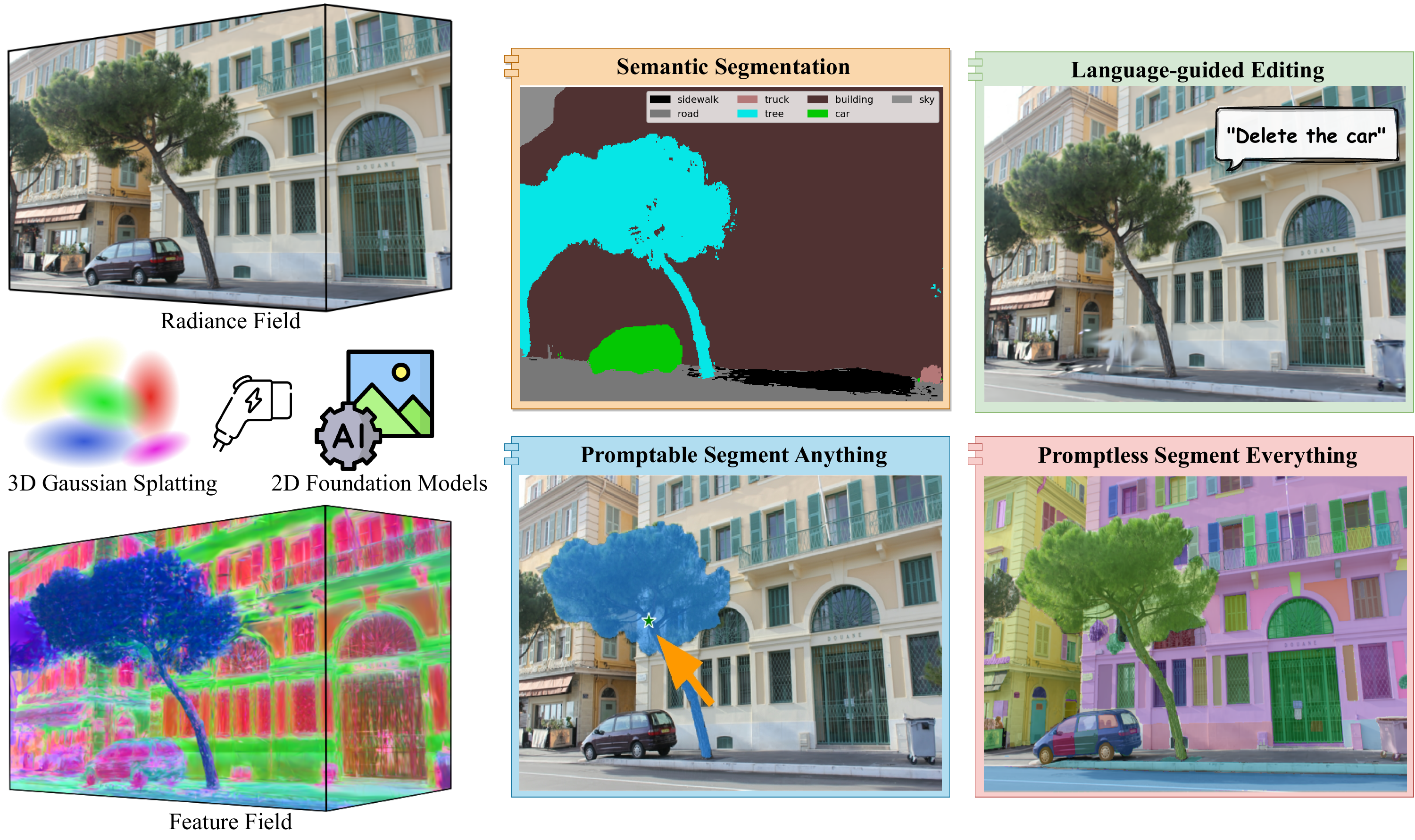} %\vspace{-2mm}
    \vspace{-5mm}
    \hfill\caption{\textbf{Feature 3DGS}. We present a general method that significantly enhances 3D Gaussian Splatting through the integration of large 2D foundation models via feature field distillation. This advancement extends the capabilities of 3D Gaussian Splatting beyond mere novel view synthesis. It now encompasses a range of functionalities, including semantic segmentation, language-guided editing, and promptable segmentations such as ``segment anything" or automatic segmentation of everything from any novel view. Scene from~\cite{hedman2018deep}.}
    \label{fig:teaser}
    \hfill \vspace{0mm}
    %\captionof{figure}{Test caption}
}]
{
\renewcommand{\thefootnote}{\fnsymbol{footnote}}
\footnotetext[1]{Equal contribution.}
}

\begin{abstract}
3D scene representations have gained immense popularity in recent years. Methods that use Neural Radiance fields are versatile for traditional tasks such as novel view synthesis. In recent times, some work has emerged that aims to extend the functionality of NeRF beyond view synthesis, for semantically aware tasks such as editing and segmentation using 3D feature field distillation from 2D foundation models. However, these methods have two major limitations: (a) they are limited by the rendering speed of NeRF pipelines, and (b) implicitly represented feature fields suffer from continuity artifacts reducing feature quality. Recently, 3D Gaussian Splatting has shown state-of-the-art performance on real-time radiance field rendering. In this work, we go one step further: in addition to radiance field rendering, we enable 3D Gaussian splatting on arbitrary-dimension semantic features via 2D foundation model distillation. This translation is not straightforward: naively incorporating feature fields in the 3DGS framework encounters significant challenges, notably the disparities in spatial resolution and channel consistency between RGB images and feature maps. We propose architectural and training changes to efficiently avert this problem. Our proposed method is general, and our experiments showcase novel view semantic segmentation, language-guided editing and segment anything through learning feature fields from state-of-the-art 2D foundation models such as SAM and CLIP-LSeg. Across experiments, our distillation method is able to provide comparable or better results, while being significantly faster to both train and render. Additionally, to the best of our knowledge, we are the first method to enable point and bounding-box prompting for radiance field manipulation, by leveraging the SAM model. Project website at: \url{https://feature-3dgs.github.io/}.
\end{abstract}    
\section{Introduction}
\label{sec:intro}

3D scene representation techniques have been at the forefront of computer vision and graphics advances in recent years. Methods such as Neural Radiance Fields (NeRFs)~\cite{mildenhall2021nerf}, and works that have followed up on it, have enabled learning implicitly represented 3D fields that are supervised on 2D images using the rendering equation. These methods have shown great promise for tasks such as novel view synthesis. However, since the implicit function is only designed to store local radiance information at every 3D location, the information contained in the field is limited from the perspective of downstream applications.

More recently, NeRF-based methods have attempted to use the 3D field to store additional descriptive features for the scene, in addition to the radiance~\cite{kobayashi2022decomposing, fan2022nerf, kerr2023lerf, tschernezki22neural}. These features, when rendered into feature images, can then provide additional semantic information for the scene, enabling donwstream tasks such as editing, segmentation and so on. However, feature field distillation through such a method is subject to a major disadvantage: NeRF-based methods can be natively slow to train as well as to infer. This is further complicated by model capacity issues: if the implicit representation network is kept fixed, while requiring it to learn an additional feature field (to not make the rendering and inference speeds even slower), the quality of the radiance field, as well as the feature field is likely to be affected unless the weight hyperparameter is meticulously tuned~\cite{kobayashi2022decomposing}.

A recent alternative for implicit radiance field representations is the 3D Gaussian splatting-based radiance field proposed by Kerbl et al.~\cite{kerbl20233d}. This explicitly-represented field using 3D Gaussians is found to have superior training speeds and rendering speeds when compared with NeRF-based methods, while retaining comparable or better quality of rendered images. This speed of rendering while retaining high quality has paved the way for real-time rendering applications, such as in VR and AR, that were previously found to be difficult. However, the 3D Gaussian splatting framework suffers the same representation limitation as NeRFs: natively, the framework does not support joint learning of semantic features and radiance field information at each Gaussian.

In this work, we present Feature 3DGS: the first feature field distillation technique based on the 3D Gaussian Splatting framework. Specifically, we propose learning a semantic feature at each 3D Gaussian, in addition to color information. Then, by splatting and rasterizing the feature vectors differentiably, the distillation of the feature field is possible using guidance from 2D foundation models. While the structure is natural and simple, enabling fast yet high-quality feature field distillation is not trivial: as the dimension of the learnt feature at each Gaussian increases, both training and rendering speeds drop drastically. We therefore propose learning a structured lower-dimensional feature field, which is later upsampled using a lightweight convolutional decoder at the end of the rasterization process. Therefore, this pipeline enables us to achieve improved feature field distillation at faster training and rendering speeds than NeRF-based methods, enabling a range of applications, including semantic segmentation, language-guided editing, promptable/promptless instance segmentation and so on.

In summary, our contributions are as follows: 
%\begin{enumerate}
\begin{itemize}
    \item A novel 3D Gaussian splatting inspired framework for feature field distillation using guidance from 2D foundation models.
    \item A general distillation framework capable of working with a variety of feature fields such as CLIP-LSeg, Segment Anything (SAM) and so on.
    \item Up to \textbf{2.7$\times$} faster feature field distillation and feature rendering over NeRF-based method by leveraging low-dimensional distillation followed by learnt convolutional upsampling.
    \item Up to \textbf{23\%} improvement on mIoU for tasks such as semantic segmentation.
%\end{enumerate}
\end{itemize}

\begin{figure*}[t]
  \centering
%   \fbox{\rule{0pt}{3in} \rule{0.9\linewidth}{0pt}}
   \includegraphics[width=\linewidth]{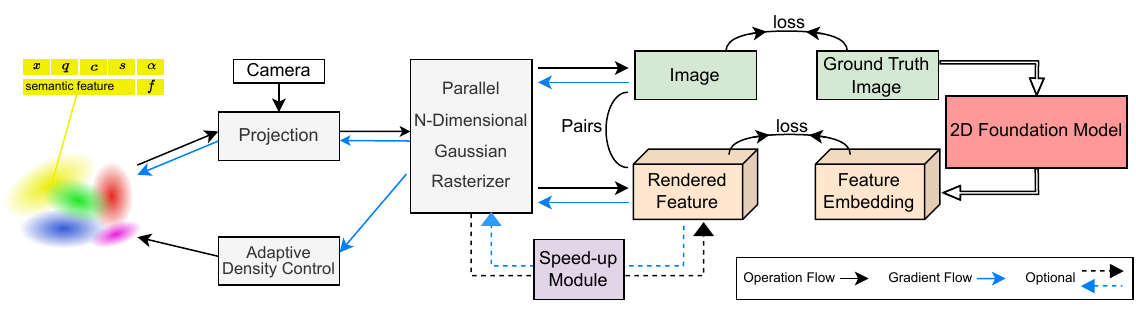}

   \caption{\textbf{An overview of our method.} We adopt the same 3D Gaussian initialization from sparse SfM point clouds as utilized in 3DGS, with the addition of an essential attribute: the \textit{semantic feature}. Our primary innovation lies in the development of a Parallel N-dimensional Gaussian Rasterizer, complemented by a convolutional speed-up module as an optional branch. This configuration is adept at rapidly rendering arbitrarily high-dimensional features without sacrificing downstream performance.}
   \label{fig:pipeline}
\end{figure*}

\section{Related Work}
\label{sec:related_works}
\subsection{Implicit Radiance Field Representations}

Implicit neural representations have achieved remarkable success in recent years across a variety of areas within the field of computer vision~\cite{park2019deepsdf, peng2020convolutional, mescheder2019occupancy,wang2023alto,mildenhall2021nerf,mipnerf}. NeRF~\cite{mildenhall2021nerf} demonstrates outstanding performance in novel view synthesis by representing 3D scenes with a coordinate-based neural network.
In mip-NeRF~\cite{mipnerf}, point-based ray tracing is replaced using cone tracing to combat aliasing. 
Zip-NeRF~\cite{zipnerf} utilized an anti-aliased grid-based technique to boost the radiance field performance. 
Instant-NGP~\cite{muller2022instant} reduces the cost for neural primitives with a versatile new input encoding that permits the use of a smaller network without sacrificing quality, thus significantly reducing the number of floating point and memory access operations.
IBRNet~\citep{wang2021ibrnet}, MVSNeRF~\citep{mvsnerf}, and PixelNeRF~\citep{pixelnerf} construct a generalizable 3D representation by leveraging features gathered from various observed viewpoints.
However, NeRF-based methods are hindered by slow rendering speeds and substantial memory usage during training, a consequence of their implicit design.

\subsection{Explicit Radiance Field Representations}
Pure implicit radiance fields are slow to operate and usually require millions of times querying a neural network for rendering a large-scale scene. Marrying explicit representations into implicit radiance fields enjoys the best of both worlds. Triplane~\cite{chan2022efficient}, TensoRF~\cite{chen2022tensorf}, K-Plane~\cite{fridovich2023k}, TILED~\cite{yi2023canonical} adopt tensor factorization to obtain efficient explicit representation. InstantNGP~\cite{muller2022instant} utilizes multi-scale hash grids to work with large-scale scenes. 
Block-NeRF~\cite{tancik2022block} further extends NeRF to render city-scale scenes spanning multiple blocks.
Point NeRF~\cite{xu2022point} uses neural 3D points for representing and rendering a continuous radiance volume.
NU-MCC~\cite{lionar2023nu} similarly utilizes latent point features but focuses on shape completion tasks.
Unlike NeRF-style volumetric rendering, 3D Gaussian Splatting introduces point-based $\alpha$-blending and an efficient point-based rasterizer.
Our work follows 3D Gaussians Splatting, where we represent the scene using explicit point-based 3D representation, \textit{i.e.} anisotropic 3D Gaussians.

\subsection{Feature Field Distillation}
Enabling simultaneously novel view synthesis and representing feature fields is well explored under NeRF~\cite{mildenhall2021nerf} literature. Pioneering works such as Semantic NeRF~\cite{semantic_nerf} and Panoptic Lifting~\cite{panoptic_lifting} have successfully embedded semantic data from segmentation networks into 3D spaces. Their research has shown that merging noisy or inconsistent 2D labels in a 3D environment can yield sharp and precise 3D segmentation. Further extending this idea, techniques like those presented in~\cite{ren-cvpr2022-nvos} have demonstrated the effectiveness of segmenting objects in 3D with minimal user input, like rudimentary foreground-background masks. Beyond optimizing NeRF with estimated labels, Distilled Feature Fields~\cite{kobayashi2022decomposing}, NeRF-SOS~\cite{fan2022nerf}, LERF~\cite{kerr2023lerf}, and Neural Feature Fusion Fields~\cite{tschernezki22neural} have delved into embedding pixel-aligned feature vectors from technologies such as LSeg or DINO~\cite{dino} into NeRF frameworks. Additionally,~\cite{mazur2023feature, tsagkas2023vl, goel2023interactive, liu2024weakly, shafiullah2022clip, shen2023distilled, ye2023featurenerf} also explore feature fusion and manipulation in 3D. Feature 3DGS shares a similar idea for distilling 2D well-trained models, but also demonstrates an effective way of distilling into explicit point-based 3D representations, for simultaneous photo-realistic view synthesis and label map rendering.

\section{Method}
\label{sec:method}

% NeRF-based feature field distillation, as explored in~\cite{kobayashi2022decomposing}, utilizes two distinct branches of MLPs to output the color $c$ and feature $f$. Subsequently, the RGB image and high-dimensional feature map are rendered individually through volumetric rendering. The transition from NeRF to 3DGS is not as straightforward as simply rasterizing RGB images and feature maps independently. Typically, feature maps have fixed dimensions that often differ from those of RGB images. Due to the shared attributes between images and feature maps, rendering them independently can be problematic (this is referred to as warp-level divergence~\cite{xiang2014warp}). A naive approach is to adopt a two-stage training method that rasterizes them separately. However, this approach could result in inoptimal quality for both RGB images and feature maps, given the high-dimensional correlations of semantic features with the shared attributes of RGB.

NeRF-based feature field distillation, as explored in~\cite{kobayashi2022decomposing}, utilizes two distinct branches of MLPs to output the color $c$ and feature $f$. Subsequently, the RGB image and high-dimensional feature map are rendered individually through volumetric rendering. The transition from NeRF to 3DGS is not as straightforward as simply rasterizing RGB images and feature maps independently. Typically, feature maps have fixed dimensions that often differ from those of RGB images. Due to the tile-based rasterization procedure and shared attributes between images and feature maps, rendering them independently can be problematic. A naive approach is to adopt a two-stage training method that rasterizes them separately. However, this approach could result in suboptimal quality for both RGB images and feature maps, given the high-dimensional correlations of semantic features with the shared attributes of RGB.

In this section, we introduce a novel pipeline for high-dimensional feature rendering and feature field distillation, which enables 3D Gaussians to explicitly represent both radiance fields and feature fields. Our proposed parallel N-dimensional Gaussian rasterizer and speed-up module can effectively solve the aforementioned problems and is capable of rendering arbitrary dimensional semantic feature map. An overview of our method is shown in~\cref{fig:pipeline}.
Our proposed method is general and compatible with any 2D foundation model, by distilling the semantic features into a 3D feature field using 3D Gaussian splatting. In our experiments, we employ SAM~\cite{kirillov2023segment} and LSeg~\cite{li2022languagedriven}, facilitating promptable, promptless (zero-shot~\cite{bucher2019zero}~\cite{gu2020context}) and language-driven computer vision tasks in a 3D context.

\subsection{High-dimensional Semantic Feature Rendering} 

% Details of high-dimensional semantic feature rendering ($\alpha$ - blending equation, CUDA parrallel computing, 3DGS + CNN design etc.)

% Why our explicit representation better than implicit representation (NeRF): in NeRF radiance field and feature field share the same neural network weights. Our joint-optimization scheme is a parallel fashion and powered by CUDA.

To develop a general feature field distillation pipeline, our method should be able to render 2D feature maps of arbitrary size and feature dimension, in order to cope with different kinds of 2D foundation models. To achieve this, we use the rendering pipeline based on the differentiable Gaussian splatting framework proposed by~\cite{kerbl20233d} as our foundation. We follow the same 3D Gaussians initialization technique using Structure from Motion~\cite{schonberger2016structure}. Given this initial point cloud, each point $x\in \mathbb{R}^{3}$ within it can be described as the center of a Gaussian. In world coordinates, the 3D Gaussians are defined by a full 3D covariance matrix $\Sigma$, which is transformed to $\Sigma'$ in camera coordinates when 3D Gaussians are projected to 2D image / feature map space~\cite{zwicker2001ewa}:
\begin{equation}
\Sigma' = JW \Sigma W^{T} J^{T}, 
\label{eq:splatting}
\end{equation}

where $W$ is the world-to-camera transformation matrix and $J$ is the Jacobian of the affine approximation of the projective transformation. $\Sigma$ is physically meaningful only when it is positive semi-definite --- a condition that cannot always be guaranteed during optimization. This issue can be addressed by decomposing $\Sigma$ into rotation matrix $R$ and scaling matrix $S$:
\begin{equation}
\Sigma = R S S^{T} R^{T}, 
\label{eq:splatting}
\end{equation}

Practically, the rotation matrix $R$ and the scaling matrix $S$ are stored as a rotation quaternion $q\in \mathbb{R}^{4}$ and a scaling factor $s\in \mathbb{R}^{3}$ respectively. Besides the aforementioned optimizable parameters, an opacity value $\alpha \in \mathbb{R}$ and spherical harmonics (SH) up to the 3rd order are also stored in the 3D Gaussians. In practice, we optimize the zeroth-order SH for the first 1000 iterations, which equates to a simple diffuse color representation $c \in \mathbb{R}^{3}$, and we introduce 1 band every 1000 iterations until all 4 bands of SH are represented. Additionally, we incorporate the semantic feature $f \in \mathbb{R}^{N}$, where $N$ can be any arbitrary number representing the latent dimension of the feature. In summary, for the $i-$th 3D Gaussian, the optimizable attributes are given by $\Theta_{i} = \{x_i, q_i, s_i, \alpha_i, c_i, f_i \}$. 

Upon projecting the 3D Gaussians into a 2D space, the color $C$ of a pixel and the feature value $F_{s}$ of a feature map pixel are computed by volumetric rendering which is performed using front-to-back depth order~\cite{kopanas2021point}:
\begin{equation}
C = \sum_{i \in \mathcal{N}} c_i \alpha_iT_i, \quad F_{s} = \sum_{i \in \mathcal{N}} f_i \alpha_iT_i, 
\label{eq:frender}
\end{equation}
where $T_i = \prod_{j=1}^{i-1} (1 - \alpha_j)$,  $\mathcal{N}$ is the set of sorted Gaussians overlapping with the given pixel, $T_i$ is the transmittance, defined as the product of opacity values of previous Gaussians overlapping the same pixel. 
The subscript $s$ in $F_{s}$ denotes ``student", indicating that this rendered feature is per-pixel supervised by the ``teacher" feature $F_{t}$. The latter represents the latent embedding obtained by encoding the ground truth image using the encoder of 2D foundation models. This supervisory relationship underscores the instructional dynamic between $F_{s}$ and $F_{t}$ in our model. In essence, our approach involves distilling~\cite{hinton2015distilling} the large 2D teacher model into our small 3D student explicit scene representation model through differentiable volumetric rendering.

In the rasterization stage, we adopted a joint optimization method, as opposed to rasterizing the RGB image and feature map independently. Both image and feature map utilize the same tile-based rasterization procedure, where the screen is divided into $16 \times 16$ tiles, and each thread processes one pixel. Subsequently, 3D Gaussians are culled against both the view frustum and each tile. Owing to their shared attributes, both the feature map and RGB image are rasterized to the same resolution but in different dimensions, corresponding to the dimensions of $c_i$ and $f_i$
initialized in the 3D Gaussians. This approach ensures that the fidelity of the feature map is rendered as high as that of the RGB image, thereby preserving per-pixel accuracy.

\subsection{Optimization and Speed-up}

The loss function is the photometric loss combined with the feature loss:
\begin{equation}
\mathcal{L} = \mathcal{L}_{rgb} + \gamma \mathcal{L}_{f},
\label{eq:loss}
\end{equation}
with
\begin{align*}
\mathcal{L}_{\text{rgb}} &= (1 - \lambda)\mathcal{L}_{1} (I,\hat{I}) + \lambda \mathcal{L}_{D-SSIM}(I,\hat{I}),\\
\mathcal{L}_{f} &= \| F_{t}(I) - F_{s}(\hat{I}) \|_1.
\end{align*}
where $I$ is the ground truth image and $\hat{I}$ is our rendered image. The latent embedding \( F_{t}(I) \) is derived from the 2D foundation model by encoding the image \( I \), while \( F_{s}(\hat{I}) \) represents our rendered feature map. To ensure identical resolution $H \times W$ for the per-pixel \( \mathcal{L}_{1} \) loss calculation, we apply bilinear interpolation to resize \( F_{s}(\hat{I}) \) accordingly. In practice, we set the weight hyperparameters $\gamma = 1.0$ and $\lambda = 0.2$.

% \subsection{Supercharging 3DGS: Semantic meaningful and editable novel view synthesis}
% Details about how we achieve the semantic segmentation using this feature field distillation technique, e.g. Lseg, SAM

% Why we have to learn a feature field instead of naively rendering a naive novel view and then apply 2D foundation models? (Speed slow, not editable and segmentable in 3D)

% Math of editable novel view synthesis, how to mathematically operate on explicit 3D gaussian. 

It is important to note that in NeRF-based feature field distillation, the scene is implicitly represented as a neural network. In this configuration, as discussed in~\cite{kobayashi2022decomposing}, the branch dedicated to the feature field shares some layers with the radiance field. This overlap could potentially lead to interference, where learning the feature fields might adversely affect the radiance fields. To address this issue, a compromise approach is to set $\gamma$ to a low value, meaning the weight of the feature field is much smaller than that of the radiance field during the optimization.~\cite{kobayashi2022decomposing} also mentions that NeRF is highly sensitive to $\gamma$. Conversely, our explicit scene representation avoids this issue. Our equal-weighted joint optimization approach has demonstrated that the resulting high-dimensional semantic features significantly contribute to scene understanding and enhance the depiction of physical scene attributes, such as opacity and relative positioning. See the comparison between Ours and Base 3DGS in~\cref{tab:dffcompare}.

To optimize the semantic feature $f \in \mathbb{R}^N$, we minimize the difference between the rendered feature map $F_s(\hat{I}) \in \mathbb{R}^{H \times W \times N}$ and the teacher feature map $F_t(I) \in \mathbb{R}^{H \times W \times M}$, ideally with $N=M$. However, in practice, $M$ tends to be a very large number due to the high latent dimensions in 2D foundation models (e.g. $M = 512$ for LSeg and $M = 256$ for SAM), making direct rendering of such high-dimensional feature maps time-consuming. To address this issue, we introduce a speed-up module at the end of the rasterization process. This module consists of a lightweight convolutional decoder that upsamples the feature channels with kernel size 1$\times$1. Consequently, it is feasible to initialize $f \in \mathbb{R}^N$ on 3D Gaussians with any arbitrary $N \ll M$ and to use this learnable decoder to match the feature channels. This allows us to not only effectively achieve $F_s(\hat{I}) \in \mathbb{R}^{H \times W \times M}$, but also significantly speed up the optimization process without compromising the performance on downstream tasks.

The advantages of implementing this convolutional speed-up module are threefold: Firstly, the input to the convolution layer, with a kernel size of \(1\times1\), is the resized rendered feature map, which is significantly smaller in size compared to the original image. This makes the \(1\times1\) convolution operation computationally efficient. Secondly, this convolution layer is a learnable component, facilitating channel-wise communication within the high-dimensional rendered feature, enhancing the feature representation. Lastly, the module's design is optional. Whether included or not, it does not impact the performance of downstream tasks, thereby maintaining the flexibility and adaptability of the entire pipeline.

\subsection{Promptable Explicit Scene Representation}
Foundation models provide a base layer of knowledge and skills that can be adapted for a variety of specific tasks and applications. We wish to use our feature field distillation approach to enable practical 3D representations of these features. Specifically, we consider two foundation models, namely Segment Anything~\cite{kirillov2023segment}, and LSeg~\cite{li2022languagedriven}.
The \textit{Segment Anything Model (SAM)} ~\cite{kirillov2023segment} allows for both promptable and promptless zero-shot segmentation in 2D, without the need for specific task training. LSeg~\cite{li2022languagedriven} introduces a language-driven approach to zero-shot semantic segmentation. Utilizing the image feature encoder with the DPT architecture~\cite{ranftl2021vision} and text encoders from CLIP~\cite{radford2021learning}, LSeg extends text-image associations to a 2D pixel-level granularity. Through the teacher-student distillation, \textbf{our distilled feature fields facilitate the extension of all 2D functionalities --- prompted by point, box, or text --- into the 3D realm.}

%Instead of naively rendering a novel view image and then applying a 2D foundation model for segmentation, we integrate the rendered feature to the decoder of the foundation model to perform segmentation tasks. Through the adoption of this method, enhanced segmentation speed can be attained, thereby facilitating a closer approximation to real-time semantic segmentation. By experiments, we have ascertained that the segmentation speed can be effectively doubled in comparison to the approach of directly applying a segmentation model to the rendered image, which is further explained in section~\ref{sec:Lseg segmentation results results}.

Our promptable explicit scene representation works as follows: for a 3D Gaussian $x$ among the $N$ ordered Gaussians overlapping the target pixel, i.e. $x_i \in \mathcal{X} $ where $\mathcal{X} =\left\{x_1, \ldots, x_N\right\}$, the activation score of a prompt $\tau$ on the 3D Gaussian $x$ is calculated by cosine similarity between the query $q(\tau)$ in the feature space and the semantic feature $f(x)$ of the 3D Gaussian followed by a softmax:

\begin{equation}
s = \frac{f(x) \cdot q(\tau)}{\|f(x)\| \|q(\tau)\|},
\end{equation}

If we have a set $\mathcal{T}$ of possible labels, such as a text label set for semantic segmentation or a point set of all the possible pixels for point-prompt, the probability of a prompt $\tau$ of a 3D Gaussian can be obtained by softmax:

\begin{equation}
\mathbf{p}(\tau | x) = \text{softmax} (s) = \frac{\exp \left( s \right)}{\sum_{s_j \in \mathcal{T}} \exp \left(s_j\right)}.
\end{equation}

%We then leverage this probability to select the Gaussians by masking out the Gaussians with low probability scores. Based on the selection, we can perform any operation such as extraction, deletion, or color changing by updating the color $c(x)$ and opacity $\alpha(x)$ accordingly. Now that we have the updated color set $\left\{c_i\right\}_{i=1}^n$ and opacity set $\left\{\alpha_i\right\}_{i=1}^n$ ($n$ is smaller than $N$), we can apply point-based $\alpha$-blending to render the edited radiance field from any view.

We utilize the computed probabilities to filter out Gaussians with low probability scores. This selective approach enables various operations, such as extraction, deletion, or appearance modification, by updating the color \( c(x) \) and opacity \( \alpha(x) \) values as needed. With the newly updated color set \( \{c_i\}_{i=1}^n \) and opacity set \( \{\alpha_i\}_{i=1}^n \), where \( n \) is smaller than \( N \), we can implement point-based \( \alpha \)-blending to render the edited radiance field from any novel view.

\begin{table}[t]
    % \vskip 0.05in
\begin{center}
\resizebox{\linewidth}{!}{
\begin{tabular}{lccc} 
\toprule
Metrics & PSNR\scalebox{0.7}{($\pm$s.d.)}$\uparrow$ & SSIM\scalebox{0.7}{($\pm$s.d.)}$\uparrow$ & LPIPS\scalebox{0.7}{($\pm$s.d.)}$\downarrow$\\
\midrule
 Ours (w/ speed-up) & \textbf{37.012} \scalebox{0.7}{($\pm$0.07)} & \textbf{0.971} \scalebox{0.7}{($\pm$5.3e-4)}& \textbf{0.023}  \scalebox{0.7}{$(\pm$2.9e-4)}\\
 Ours & 36.915 \scalebox{0.7}{($\pm$0.05)} & 0.970 \scalebox{0.7}{($\pm$5.7e-4)}& 0.024  \scalebox{0.7}{($\pm$1.1e-3)}\\
 Base 3DGS & 36.133 \scalebox{0.7}{($\pm$0.06)}& 0.965 \scalebox{0.7}{($\pm$1.5e-4)}& 0.033 \scalebox{0.7}{($\pm$1.2e-3)}\\
\bottomrule
\end{tabular}}\vspace{-3mm}
\end{center}
\caption{\textbf{Performance on Replica Dataset.} (average performance for 5K training iterations, speed-up module rendered feature $dim = 128$). Boldface font represents the preferred results.}
\label{tab:dffcompare}
\end{table}
%%%%%%%%%%%%%%%%%%%%%%%%%%%%%%%%%%%%%%%%

\begin{table}[t]
\centering
% \resizebox{0.5\textwidth}{!}{ % Table spans the full text width
\begin{tabular}{lccc}
\toprule
 Metrics &  mIoU$\uparrow$ & accuracy$\uparrow$ & FPS$\uparrow$ \\
\midrule
Ours (w/ speed-up)  & 0.782 & \textbf{0.943} & \textbf{14.55} \\
Ours  & \textbf{0.787}&\textbf{0.943} & 6.84 \\
NeRF-DFF & 0.636& 0.864 & 5.38\\
\bottomrule
\end{tabular}
% }
\caption{\textbf{Performance of semantic segmentation on Replica dataset compared to NeRF-DFF.} (speed-up module rendered feature $dim = 128$). Boldface font represents the preferred results.}
\label{table:replicaperform}
\end{table}
\section{Experiments}
\label{sec:experiments}

\begin{figure*}[t]
%   \fbox{\rule{0pt}{3in} \rule{0.9\linewidth}{0pt}}

  \includegraphics[width=\linewidth]{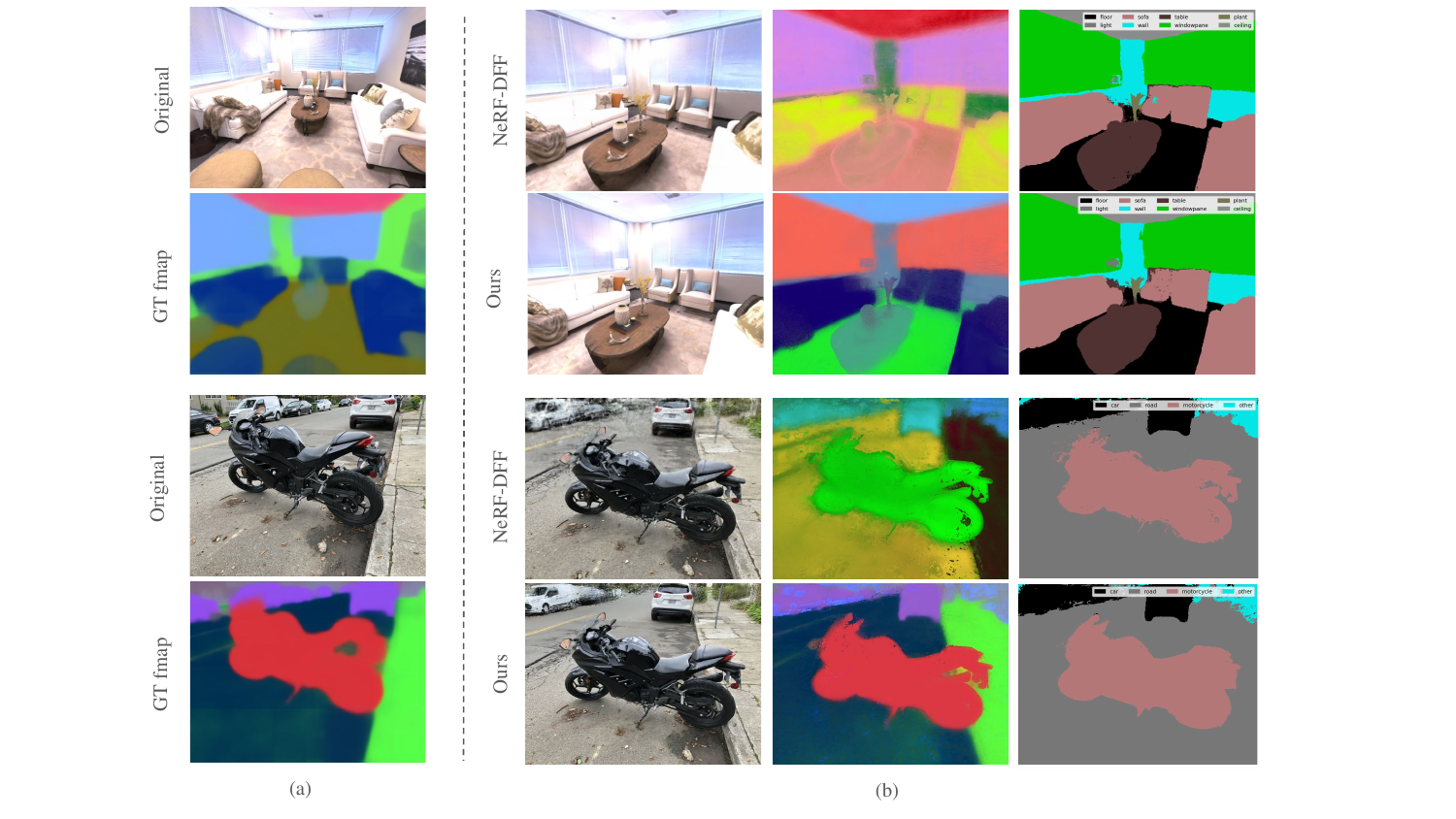}\vspace{-3mm}

  \caption{\textbf{Novel view semantic segmentation (LSeg) results on scenes from Replica dataset~\cite{straub2019replica} and LLFF dataset~\cite{mildenhall2019local}.} (a) We show examples of original images in training views together with the ground-truth feature visualizations. (b) We compare the qualitative segmentation results using our Feature 3DGS with the NeRF-DFF~\cite{kobayashi2022decomposing}. Our inference is \textbf{1.66}$\times$ faster when rendered feature $dim = 128$. Our method demonstrates more fine-grained segmentation results with higher-quality feature maps.} 
  \label{fig:results1_lseg_segmentation}
\end{figure*}

\subsection{Novel view semantic segmentation}
\label{subsec:novel view semantic segmentation}

The number of classes of a dataset is usually limited from tens~\cite{everingham2015pascal} to hundreds~\cite{zhou2019semantic}, which is insignificant to English words~\cite{li2020fss}. In light of the limitation, semantic features empower models to comprehend unseen labels by mapping semantically close labels to similar regions in the embedding space, as articulated by Li et al~\cite{li2022languagedriven}. This advancement notably promotes the scalability in information acquisition and scene understanding, facilitating a profound comprehension of intricate scenes. We distill LSeg feature for this novel view semantic segmentation task. Our experiments demonstrate the improvement of incorporating semantic feature over the naive 3D Gaussian rasterization method~\cite{kerbl20233d}. In~\cref{tab:dffcompare}, we show that our model surpasses the baseline 3D Gaussian model in performance metrics on Replica dataset~\cite{straub2019replica} with 5000 training iterations for all three models. Noticeably, the integration of the speed-up module to our model does not compromise the performance.

In our further comparison with NeRF-DFF~\cite{kobayashi2022decomposing} using the Replica dataset, we address the potential trade-off between the quality of the semantic feature map and RGB images. In~\cref{table:replicaperform}, our model demonstrates higher accuracy and mean intersection-over-union (mIoU). Additionally, by incorporating our speed-up module, we achieved more than double the frame rate per second (FPS) of our full model while maintaining comparable performance. In~\cref{fig:results1_lseg_segmentation} (b) the last column, our approach yields better visual quality on novel views and semantic segmentation masks for both synthetic and real scenes compared to NeRF-DFF.

\begin{figure}[t]
%   \fbox{\rule{0pt}{3in} \rule{\linewidth}{0pt}}
 \includegraphics[width=\linewidth]{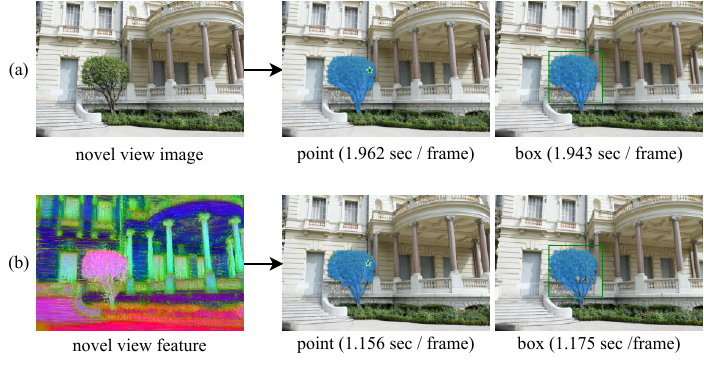}\vspace{-2mm}      
     \vspace{-.2cm}\caption{\textbf{Comparison of SAM segmentation results obtained by} (a) naively applying the SAM encoder-decoder module to a novel-view rendered image \textbf{with} (b) directly decoding a rendered feature. Our method is up to $1.7\times$ faster in total inference speed including rendering and segmentation while preserving the quality of segmentation masks. Scene from~\cite{hedman2018deep}.}
  \label{fig:results2_sam_segementation_fea_img}
\end{figure}

\begin{figure*}[t]
%   \fbox{\rule{0pt}{6in} \rule{0.9\linewidth}{0pt}}
   \includegraphics[width=\linewidth]{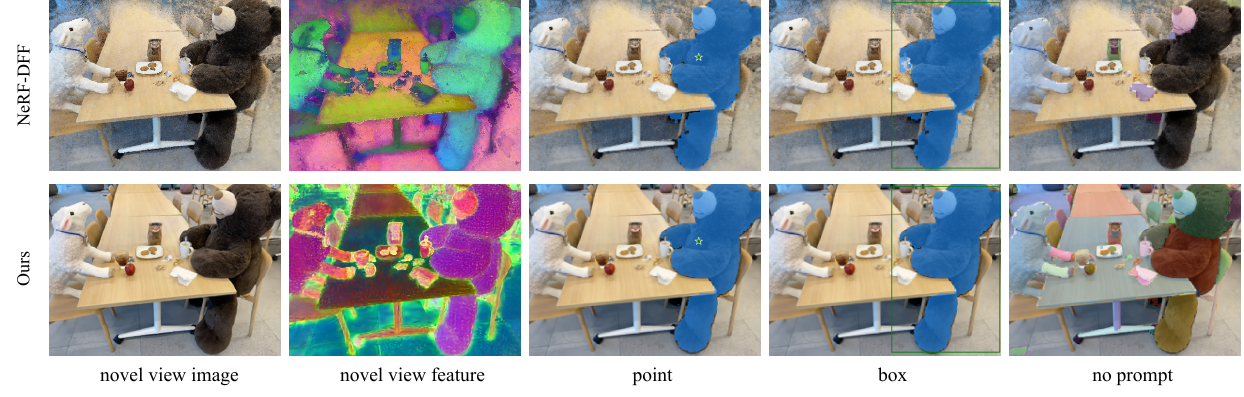}\vspace{-2mm}
\caption{\textbf{Novel view segmentation (SAM) results compared with NeRF-DFF}. (Upper) 
NeRF-DFF method presents lower-quality segmentation masks - note the failure on segmenting the cup from the bear and the coarse-grained mask boundary on the bear's leg in box-prompted results. (Lower) Our method provides higher-quality masks with more fine-grained segmentation details. Scene from~\cite{kerr2023lerf}.}
  \label{fig:results2_sam_segementation_nerf_ours}
\end{figure*}

\subsection{Segment Anything from Any View}
SAM excels in performing precise instance segmentation, utilizing interactive points and boxes as prompts to automatically segment objects in any 2D image. In our experiments, we extend this capability to 3D, aiming to achieve fast and accurate segmentation from any viewpoint. Our distilled feature field enables the model to render the SAM feature map directly for any given camera pose. As such, the SAM decoder is the only component needed to interact with the input prompt and produce the segmentation mask, thereby bypassing the need to synthesize a novel view image first and then process it through the entire SAM encoder-decoder pipeline. Furthermore, to enhance training and inference speed, we use the speed-up module in this experiment. In practice, we set the rendered feature dimension to 128, which is half of SAM's latent dimension of 256, maintaining the comparable quality of segmentation.

In~\cref{fig:results2_sam_segementation_fea_img}, we compare the results of both point and box prompted segmentation on novel views using the naive approach (SAM encoder + decoder) and our proposed feature field approach (SAM decoder only). We achieve nearly equivalent segmentation quality, but our method is up to 1.7$\times$ faster. In ~\cref{fig:results2_sam_segementation_nerf_ours}, we contrast our method with NeRF-DFF~\cite{kobayashi2022decomposing}. Our rendered features not only yield higher quality mask boundaries, as evidenced by the bear's leg, but also deliver more accurate and comprehensive instance segmentation, capable of segmenting finer-grained instances (as illustrated by the more 'detailed' mask on the far right). Additionally, we use PCA-based feature visualization~\cite{pedregosa2011scikit} to demonstrate that our high-quality segmentation masks result from superior feature rendering.

\begin{figure*}[t]
%   \fbox{\rule{0pt}{3in} \rule{0.9\linewidth}{0pt}}

  \includegraphics[width=\linewidth]{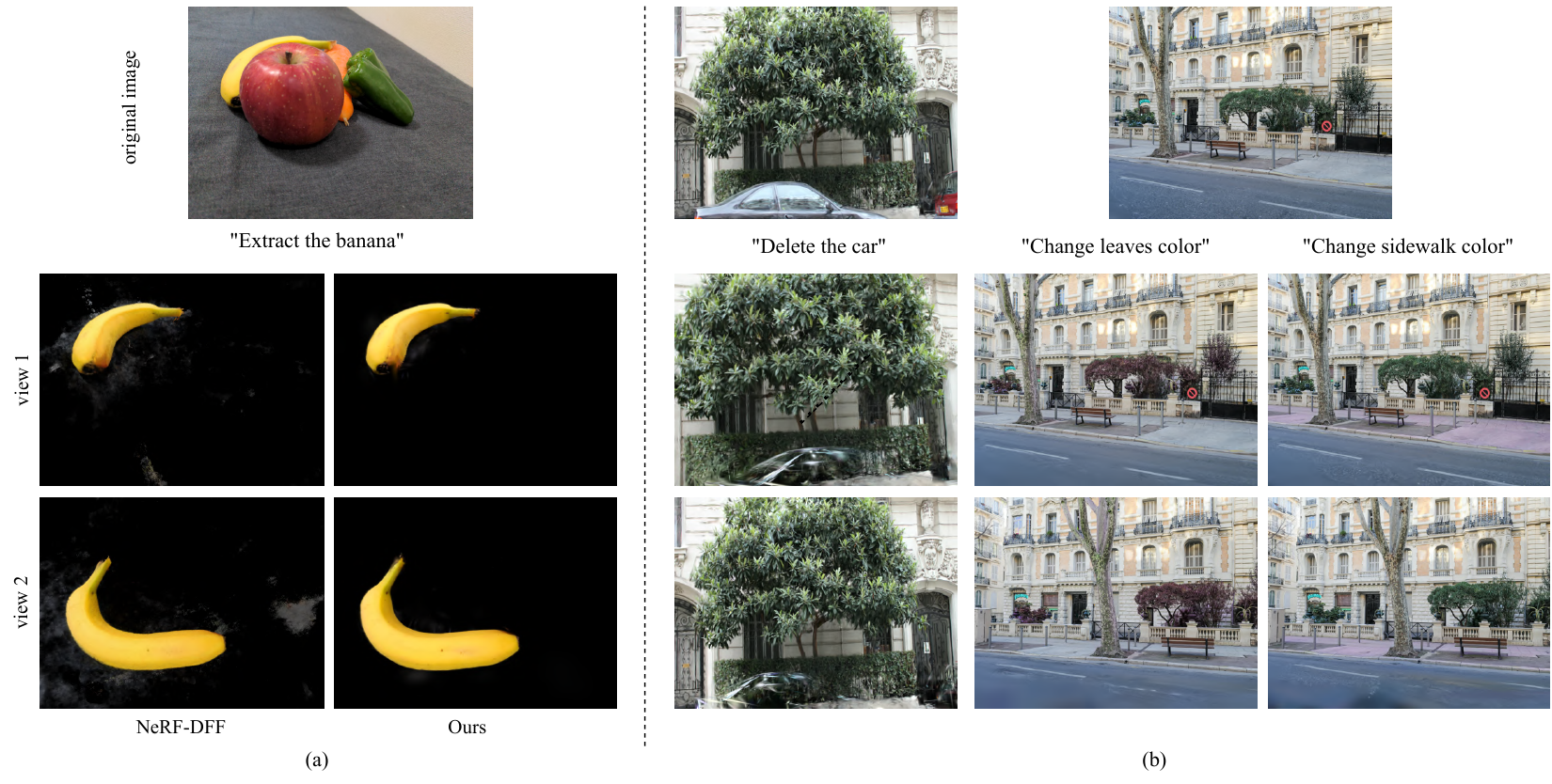}\vspace{-3mm}
  \caption{\textbf{Demonstration of results with various language-guided edit operations by querying the 3D feature field and comparison with NeRF-DFF} (a) We compare our edit results with NeRF-DFF method on the sample dataset provided by NeRF-DFF~\cite{kobayashi2022decomposing}. Note that our method outperforms NeRF-DFF method by extracting the entire banana hidden by an apple in the original image and with less floaters in the background. (b) We demonstrate results with deletion and appearance modification on different targets. Note that the car is deleted with background preserved, and the appearance of the leaves changes with the appearance of the stop sign remained the same.} 
  \label{fig:results3_lseg_edting_nerf}
\end{figure*}

% \begin{figure}[t]
% %   \fbox{\rule{0pt}{3in} \rule{\linewidth}{0pt}}
%  \includegraphics[width=\linewidth]{figures/results3_lseg_edting.pdf}\vspace{-2mm}      
%      \vspace{-.2cm}\caption{\textbf{Object-level comparisons on ShapeNet.} On the car, ALTO recovers the detail of having both side mirrors.}
%   \label{fig:results3_lseg_edting}
% \end{figure}

\subsection{Language-guided Editing}
In this section, we showcase the capability of our Feature 3DGS, distilled from LSeg, to perform editable novel view synthesis. The process begins by querying the feature field with a text prompt, typically comprising an edit operation followed by a target object, such as ``extract the car". For text encoding, we employ a ViT-B/32 CLIP encoder. Our editing pipeline capitalizes on the semantic features queried from the 3D feature field, rather than relying on a 2D rendered feature map. We compute semantic scores for each 3D Gaussian, represented by a $K$-dimensional vector (where $K$ is the number of object categories), using a softmax function. Subsequently, we engage in either soft selection (by setting a threshold value) or hard selection (by filtering based on the highest score across $K$ categories). To identify the region for editing, we generate a ``Gaussian mask" through thresholding on the score matrix, which is then utilized for modifications to color $c$ and opacity $\alpha$ on 3D Gaussians.

%In this section, we demonstrate our Feature 3DGS has the capability of performing editable novel view synthesis, distilled from LSeg. We query the feature field with a text prompt such as including an edit operation followed by a target object such as "extract the car". We then use a ViT-B/32 CLIP encoder for text encoding. In our editing pipeline, we leverage the semantic feature queried from the 3D feature field represented by 3D Gaussians rather than 2D image feature map so that the editing is insusceptible to any flaw or damage in the 2D rendered feature map. We calculate the semantic scores for each 3D Gaussian represented by a $K$-dimensional vector where $K$ is the number of object categories by applying a softmax function and perform either soft selection (setting a threshold value) or hard selection (filtering by the maximum score among $K$ categoris). We create a 'Gaussian mask' by applying thresholding on the score matrix to select the editing Gaussian region which is further used for modifying $c(x)$ and $\alpha(x)$.

%In ~\cref{fig:results3_lseg_edting}, we show our novel view editing results with various editing operations  prompts. Specifically, we perform the following edit operations, extraction, deletion, and color changing, to text-specified targets on different scenes, which are from sample fruit dataset provided by~\cite{kobayashi2022decomposing} and Deep Blending dataset~\cite{hedman2018deep}. 

In~\cref{fig:results3_lseg_edting_nerf}, we showcase our novel view editing results, achieved through various operations prompted by language inputs. Specifically, we conduct editing tasks such as extraction, deletion, and appearance modification on text-specified targets within diverse scenes. As illustrated in ~\cref{fig:results3_lseg_edting_nerf} (a), we successfully extract an entire banana from the scene. Notably, by leveraging 3D Gaussians to update the rendering parameters, our Feature 3DGS model gains an understanding of the 3D scene environment from any viewpoint. This enables the model to reconstruct occluded or invisible parts of the scene, as evidenced by the complete extraction of a banana initially hidden by an apple in view 1. Furthermore, compared with our edit results with NeRF-DFF, our method stands out by providing a cleaner extraction with little floaters in the background. Additionally, in ~\cref{fig:results3_lseg_edting_nerf} (b), our model is able to delete objects like cars while retaining the background elements, such as plants, due to the opacity updates in 3DGS, showcasing its 3D scene awareness. Moreover, we also demonstrate the model's capability in modifying the appearance of specific objects, like `sidewalk' and `leaves', without affecting adjacent objects' appearance (e.g., the `stop sign' remains red).

\section{Discussion and Conclusion}
\label{sec:discussion}
%In conclusion, our research bridges the gap between 2D foundation models and 3D Gaussian Splatting, opening the door to a brand new semantic, editable, and promptable explicit 3D scene representation. 
In this work, we present a notable advancement in explicit 3D scene representation by integrating 3D Gaussian Splatting with feature field distillation from 2D foundation models, a development that not only broadens the scope of radiance fields beyond traditional uses but also addresses key limitations of previous NeRF-based methods in implicitly represented feature fields. Our work, as showcased in various experiments including complex semantic tasks like editing, segmentation, and language-prompted interactions with models like CLIP-LSeg and SAM, opening the door to a brand new semantic, editable, and promptable explicit 3D scene representation.

However, our Feature 3DGS framework does have its inherent limitations. The student feature's limited access to the ground truth feature restricts the overall performance, and the imperfections of the teacher network further constrain our framework's effectiveness. In addition, our adaptation of the original 3DGS pipeline, which inherently generates noise-inducing floaters, poses another challenge, affecting our model's optimal performance. %Lastly, while the original 3DGS supports real-time rendering, the integration of semantic features in our model introduces a bottleneck in rendering speed, a trade-off that remains unresolved in our current framework.

\section*{Acknowledgement}
\label{sec:acknowledgement}
We thank the Visual Machines Group (VMG) at UCLA and Visual Informatics Group at UT Austin (VITA) for feedback and support. This project was supported by the US DoD LUCI (Laboratory University Collaboration Initiative) Fellowship and partially supported by ARL grants W911NF-20-2-0158 and W911NF-21-2-0104 under the cooperative A2I2 program. Z.W. is partially supported by the ARL grant W911NF2120064 under the cooperative A2I2 program, and an Army Young Investigator Award. A.K. is supported by a DARPA Young Faculty Award, NSF CAREER Award, and Army Young Investigator Award.

\clearpage
{
    \small
    \bibliographystyle{ieeenat_fullname}
    \bibliography{main}
}

\clearpage
\setcounter{section}{0}
\setcounter{figure}{0}
\setcounter{table}{0}
\maketitlesupplementary

% \section{Rationale}
% \label{sec:rationale}
% % 
% Having the supplementary compiled together with the main paper means that:
% % 
% \begin{itemize}
% \item The supplementary can back-reference sections of the main paper, for example, we can refer to \cref{sec:intro};
% \item The main paper can forward reference sub-sections within the supplementary explicitly (e.g. referring to a particular experiment); 
% \item When submitted to arXiv, the supplementary will already included at the end of the paper.
% \end{itemize}
% % 
% To split the supplementary pages from the main paper, you can use \href{https://support.apple.com/en-ca/guide/preview/prvw11793/mac#:~:text=Delete%20a%20page%20from%20a,or%20choose%20Edit%20%3E%20Delete).}{Preview (on macOS)}, \href{https://www.adobe.com/acrobat/how-to/delete-pages-from-pdf.html#:~:text=Choose%20%E2%80%9CTools%E2%80%9D%20%3E%20%E2%80%9COrganize,or%20pages%20from%20the%20file.}{Adobe Acrobat} (on all OSs), as well as \href{https://superuser.com/questions/517986/is-it-possible-to-delete-some-pages-of-a-pdf-document}{command line tools}.

% \newcommand{\achuta}[1]{\textcolor{magenta}{Achuta: #1}} 
% \newcommand{\shijie}[1]{\textcolor{orange}{Shijie: #1}} 
\newcommand{\hc}[1]{\textcolor{red}{hc: #1}}

\renewcommand\thesection{\Alph{section}} % section with alphabet
\renewcommand\thesubsection{\thesection.\arabic{subsection}} % subsection with alphabet
\renewcommand\thefigure{\Alph{figure}} % figure with alphabet
\renewcommand\thetable{\Alph{table}} % table with alphabet

% Support for easy cross-referencing
% \usepackage[capitalize]{cleveref}
\crefname{section}{Sec.}{Secs.}
\Crefname{section}{Section}{Sections}
\Crefname{table}{Table}{Tables}
\crefname{table}{Tab.}{Tabs.}

\newcommand{\tabnohref}[1]{Tab.~{\color{red}#1}} % refer to table without hyper reference
\newcommand{\fignohref}[1]{Fig.~{\color{red}#1}} % refer to figure without hyper reference
\newcommand{\secnohref}[1]{Sec.~{\color{red}#1}} % refer to section without hyper reference
\newcommand{\cnohref}[1]{[{\color{green}#1}]} % refer to citation without
\newcommand{\linenohref}[1]{Line~{\color{red}#1}}

\noindent This supplement is organized as follows:
\begin{itemize}[itemsep=0em]
    \item Section~\ref{sec:Details of Architectures} contains network architecture details;
    \item Section~\ref{sec:Training and Inference Details} contains more details on the training and inference settings;
    \item Section~\ref{sec: Teacher Features} contains the details of the teacher features from 2D foundation models;
    \item Section~\ref{sec:Replica Dataset Experiment} contains more details of Replica dataset experiment;
    \item Section~\ref{sec:Editing Algorithm and Details} contains the algorithmic details of language-guided editing;
    \item Section~\ref{sec:Ablation Studies} contains ablation studies of our method;
    \item Section~\ref{sec:Failure Cases} contains failure cases of complex scenes and reasoning analysis.
    %\item Section~\ref{sec:code} contains the code link of the comparison baselines; and
    %\item Section~\ref{sec:limitation} contains discussion on the limitation of our method and future work. %; and
%    \item Section~\ref{sec:impact} contains discussion on potential negative social impact.
    
\end{itemize}

\begin{algorithm}[h]
    \caption{Parallel N-Dimensional Gaussian Rasterization}
    \begin{algorithmic}
        \State  ${PointCloud} \gets \text{Structure from Motion}$ \Comment{Point Cloud}
        \State  ${X}, {C} \gets {PointCloud}$  \Comment{Position, Colors}
        \State  ${\Sigma}, {A}, {F} \gets {InitAttributes()}$ 
        \State  \Comment{Covariances, Opacities, Semantic Features}
        \State  ${F_t(I)} \gets {I}  \text{ applying Foundation Model}$ \Comment{Feature Map}
        \State  $i \gets 0$ \Comment{Iteration Counter}
        \Repeat
            \State ${V}, {I}, {F_t} \gets \text{GetTrainingView()}$ 
            \State  \Comment{Camera Pose, Image, Feature Map}
            \State $\hat{{I}}, {F_s} \gets \text{ParallelRasterizer}({X}, {C}, {\Sigma}, {A}, {F}, {V})$ 
            \State  \Comment{Rasterization}
            \State ${L} \gets \text{Loss}({I}, \hat{{I}}) + \lambda\text{Loss}({F_t}, {F_s})$
            \State  \Comment{Loss Calculation}
            \State ${X}, {\Sigma}, {C}, {A}, {F} \gets \text{Adam}({L})$ 
            \State \Comment{Backpropagation and Step}
            \If{IsRefinementStep$(i)$}
                \For{Gaussians$(x, q, c, \alpha, f)$}
                    \If{$\alpha < \varepsilon$ \text{or} $\text{IsTooLarge}(x, q)$}
                    \State $RemoveGaussian()$
                    \EndIf
                    \If{$\nabla_pL > \tau_p$}
                        \If{$\|{S}\| > \tau_S$} 
                            \State $SplitGaussian(x, q, c, \alpha, f)$
                            \State  \Comment{Over-reconstruction}
                        \Else
                            \State $CloneGaussian(x, q, c, \alpha, f)$
                            \State  \Comment{Under-reconstruction}
                        \EndIf
                    \EndIf
                \EndFor
            \EndIf
            \State $i \gets i + 1$ \Comment{Counter Increment}
        \Until{Convergence}
    \end{algorithmic}
    \label{alg:N-dim rasterization}
\end{algorithm}

\begin{table*}[t]
    % \vskip 0.05in
\begin{center}
\fontsize{6pt}{6pt}\selectfont
\resizebox{0.9\linewidth}{!}{
\begin{tabular}{lccccccc}
\toprule
Dimension & 8 & 16 & 32 & 64 & 128 & 256 & 512 \\
\midrule
Time & 6:40 & 7:21 & 8:51 & 12:10 & 19:55 & 48:39 & 1:29:42 \\
mIoU$\uparrow$ & 0.354 & 0.493 & 0.709 & 0.774 & 0.783 & \textbf{0.791} & 0.790 \\
Accuracy$\uparrow$ & 0.735 & 0.880 & 0.927 & 0.939 & \textbf{0.944} & \textbf{0.944} & 0.943 \\
\bottomrule
\end{tabular}}
\vspace{-3mm}
\end{center}
\caption{\textbf{Evaluation of Semantic Segmentation Performance Across Different Dimensions.} This table presents the Time, mIoU, and Accuracy corresponding to each dimension level with LSeg feature.}
\label{tab:performance_metrics}
\end{table*}

\begin{table*}[t]
    % \vskip 0.05in
\begin{center}
\fontsize{8pt}{7pt}\selectfont
\resizebox{0.9\linewidth}{!}{
\begin{tabular}{lccccccc} \toprule
Dimension & 8 & 16 & 32 & 64 & 128 & 256 & 512 \\
\midrule
 PSNR$\uparrow$ & 36.8879 & 36.8871 & 36.9671 & 36.9397 & \textbf{37.012} & 36.9474 & 36.9150 \\
 SSIM$\uparrow$ & 0.9699 & 0.9703 & 0.9706 & \textbf{0.9708} & 0.9706 & 0.9704 & 0.9703 \\
 LPIPS$\downarrow$ & 0.0234 & 0.0230 & \textbf{0.0226} & 0.0229 & 0.0228 & 0.0230 & 0.0236 \\
\bottomrule
\end{tabular}}
\vspace{-3mm}
\end{center}
\caption{\textbf{Evaluation of Image Quality Metrics Across Different Dimensions of Lseg feature.} This table presents the PSNR, SSIM, and LPIPS values corresponding to each dimension level with LSeg feature.}
\label{tab:image_quality_metrics_lseg}
\end{table*}

\vspace{-3mm}
\section{Details of Architectures}
\label{sec:Details of Architectures}
\paragraph{Parallel N-Dimensional Gaussian Rasterizer}
The parallel N-dimensional rasterizer maintains an architecture akin to the original 3DGS rasterizer. Moreover, it employs a point-based $\alpha$-blending technique for rasterizing the feature map. To mitigate the issue of inconsistent spatial resolution inherent in tile-based rasterization, we ensure that both the RGB image and the feature map are rendered at matching sizes. Additionally, the parallel N-dimensional rasterizer is adaptable to various foundational models, implying that its dimensions are flexible and can vary accordingly. The detail of the Parallel N-Dimensional Gaussian Rasterization is in~\cref{alg:N-dim rasterization}.
\paragraph{Speed-up Module}
The primary objective of our Speed-up Module is to modify the feature map channels, enabling the relatively low-dimensional semantic features rendered from 3D Gaussians to align with the high-dimensional ground truth 2D feature map. To facilitate this, we employ a convolutional layer equipped with a 1 $\times$ 1 kernel, offering a direct and efficient solution. Given that we already possess the ground truth feature map from the teacher network, which serves as a target for the rendered feature map approximation, there is no necessity for a complex CNN architecture. This approach simplifies the process, ensuring effective feature alignment without the need for intricate feature extraction mechanisms. More experimental results regarding performance of the Speed-up Module are included in~\cref{sec:Ablation Studies}.

% \begin{table*}[t]
%     % \vskip 0.05in
% \begin{center}
% \fontsize{6pt}{6pt}\selectfont
% \resizebox{0.9\linewidth}{!}{
% \begin{tabular}{lccccccc}
% \toprule
% Dimension & 8 & 16 & 32 & 64 & 128 & 256 & 512 \\
% \midrule
% Time & 6:40 & 7:21 & 8:51 & 12:10 & 19:55 & 48:39 & 1:29:42 \\
% mIoU$\uparrow$ & 0.354 & 0.493 & 0.709 & 0.774 & 0.783 & \textbf{0.791} & 0.790 \\
% Accuracy$\uparrow$ & 0.735 & 0.880 & 0.927 & 0.939 & \textbf{0.944} & \textbf{0.944} & 0.943 \\
% \bottomrule
% \end{tabular}}
% \vspace{-3mm}
% \end{center}
% \caption{\textbf{Evaluation of Semantic Segmentation Performance Across Different Dimensions.} This table presents the Time, mIoU, and Accuracy corresponding to each dimension level with LSeg features.}
% \label{tab:performance_metrics}
% \end{table*}

% \begin{table*}[t]
%     % \vskip 0.05in
% \begin{center}
% \fontsize{8pt}{7pt}\selectfont
% \resizebox{0.9\linewidth}{!}{
% \begin{tabular}{lccccccc} \toprule
% Dimension & 8 & 16 & 32 & 64 & 128 & 256 & 512 \\
% \midrule
%  PSNR$\uparrow$ & 36.8879 & 36.8871 & 36.9671 & 36.9397 & \textbf{37.012} & 36.9474 & 36.9150 \\
%  SSIM$\uparrow$ & 0.9699 & 0.9703 & 0.9706 & \textbf{0.9708} & 0.9706 & 0.9704 & 0.9703 \\
%  LPIPS$\downarrow$ & 0.0234 & 0.0230 & \textbf{0.0226} & 0.0229 & 0.0228 & 0.0230 & 0.0236 \\
% \bottomrule
% \end{tabular}}
% \vspace{-3mm}
% \end{center}
% \caption{\textbf{Evaluation of Image Quality Metrics Across Different Dimensions.} This table presents the PSNR, SSIM, and LPIPS values corresponding to each dimension level with LSeg features.}
% \label{tab:image_quality_metrics_lseg}
% \end{table*}

\section{Training and Inference Details}
\label{sec:Training and Inference Details}
For the training and inference pipeline, one option is to directly render a feature map with the dimension same as the ground-truth feature (512 for LSeg encoding and 256 for SAM encoding). Since rendering with such large dimension slows down the training, another option is to use our speed-up module: rendering a lower-dimensional feature map, which is later upsampled to the ground-truth feature dimension by a lightweight convolutional decoder. Similar to 3DGS~\cite{kerbl20233d}, we use Adam optimizer for for optimization during training and use a standard exponential decay scheduling similar to~\cite{fridovich2022plenoxels}. For image rendering, we mainly follow the 3DGS optimization strategy by using a 4 times lower image resolution and upsampling twice after 250 and 500 iterations. For feature rendering, we use Adam optimizer with a learning rate of $1e-3$. For the feature decoder network in the additional Speed-up Module, we use a separate Adam optimizer with a learning rate of $1e-4$.

\section{Teacher Features}
\label{sec: Teacher Features}
\paragraph{LSeg Feature}  For LSeg, we use CLIP ViT-L/16 image encoder for ground-truth feature preparation and and ViT-L/16 text encoder for text encoding. The ground truth feature from the LSeg image encoder has feature size $360 \times 480$ with feature dimension $512$. One can either choose to directly render a $h \times w$ feature with dimension $512$ or use the Speed-up Module by rendering a lower-dimensional feature which is later upsampled back. In practice, we use rendered feature $dim=128$ for Sec. \textcolor{red}{4.1} in our main paper.

To predict the semantic segmentation mask during inference, we reshape the rendered feature with shape (512, 360, 480) to (360 $\times$ 480, 512), referred as the image feature. The text feature from the CLIP text encoder has shape $(C, 512)$ where $C$ is the number of categories. We then apply matrix multiplication between the two to align pixel-level features and a text query feature and perform semantic segmentation using LSeg spatial regularization blocks.

%%%%%
\begin{table*}[t]
    % \vskip 0.05in
\begin{center}
\fontsize{8pt}{7pt}\selectfont
\resizebox{0.85\linewidth}{!}{
\begin{tabular}{lcccccc} \toprule
Dimension & 8 & 16 & 32 & 64 & 128 & 256 \\
\midrule
PSNR$\uparrow$  & 36.7061 & 36.9145 & 36.8793 & \textbf{36.9208} & 36.7815 & 36.9139 \\
SSIM$\uparrow$  & 0.9697 & 0.9706 & 0.9700 & 0.9701 & 0.9700 & \textbf{0.9709} \\
LPIPS$\downarrow$ & 0.0234 & 0.0229 & 0.0229 & \textbf{0.0228} & \textbf{0.0228} & 0.0230 \\
FPS$\uparrow$ & 64.7 & 53.9 & 52.7 & 32.8 & 24.2 & 8.3 \\
\bottomrule
\end{tabular}}
\vspace{-3mm}
\end{center}
\caption{\textbf{Evaluation of Image Quality Metrics Across Different Dimensions of SAM feature.} This table presents the PSNR, SSIM, LPIPS, and FPS values corresponding to each dimension level with SAM feature.}
\label{tab:image_quality_metrics_sam}
\end{table*}
%%%%%

\paragraph{SAM Feature} Following the image encoding details in SAM~\cite{kirillov2023segment}, we use an MAE~\cite{he2022masked} pre-trained ViT-H/16~\cite{dosovitskiy2020image} with 14 $\times$ 14 windowed attention and four equally-spaced global attention blocks. The SAM encoder first obtains the image resolution of 1024 $\times$ 1024 by resizing the image and padding the shorter side. The resolution is then 16 $\times$ downscaled to 64$\times$64. Since only a portion of the $64\times64$ feature map contains semantic information due to the padding operation, we crop out one side of the feature map corresponding to the longer side of the original image. Specifically, suppose the original image has the resolution of $H \times W$ where $W > H$, we crop the 64 $\times$64 feature map from SAM encoder so that the new feature resolution becomes $64W/H \times 64$ with the feature dimension of 256 corresponding to the output dimension of the SAM encoder. In practice, we use the Speed-up Module with the rendered feature $dim=128$.

To obtain the results of promptable or promptless segmentation during the inference, we perform the padding operation on the rendered feature to convert from $64W/H \times 64$ back to $64 \times 64$ so that the SAM decoder receive the equivalent semantic information as from the original SAM encoder.

\begin{figure*}[t]
%   \fbox{\rule{0pt}{3in} \rule{0.9\linewidth}{0pt}}

  \includegraphics[width=\linewidth]{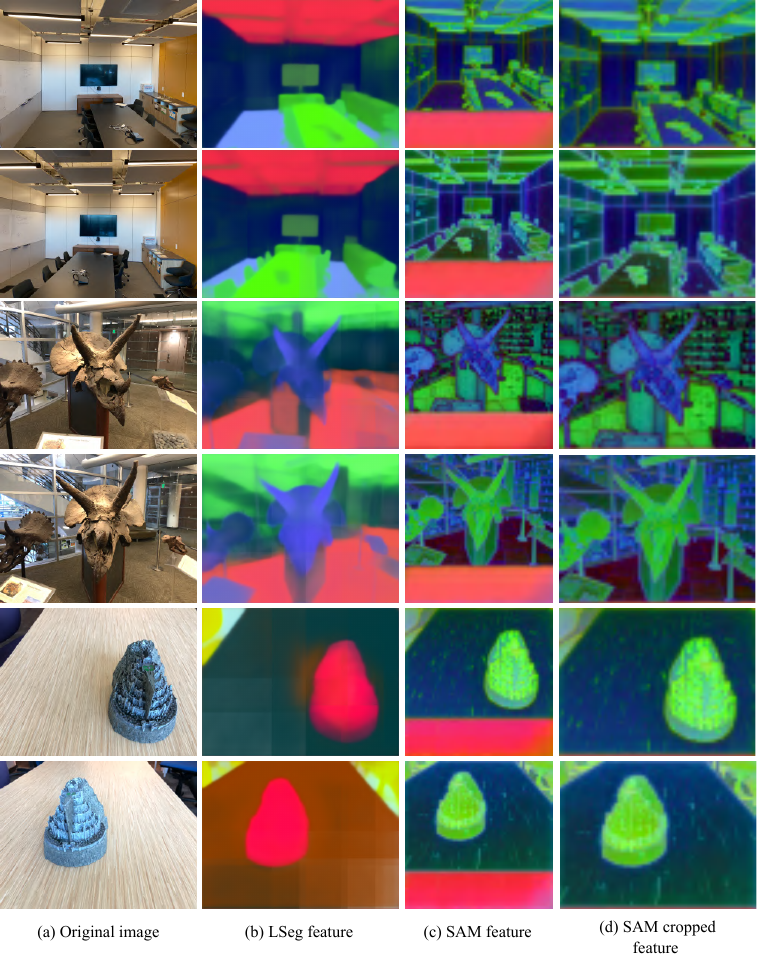}\vspace{-3mm}
  \caption{\textbf{Feature visualization on different scenes from LLFF dataset~\cite{mildenhall2019local} from LSeg and SAM encoders}. Note that SAM features in column (d) is obtained by cropping the padding region. We resize the cropped feature for better visualization.} 
  \label{fig:feature_visualization}
\end{figure*}

\paragraph{Feature visualization} As shown in ~\cref{fig:feature_visualization}, similar to ~\cite{stelzner2021decomposing}, we use \texttt{\small sklearn.decomposition.PCA}~\cite{pedregosa2011scikit} in scikit-learn package for feature visualization. We set the number of PCA components to 3 corresponding to RGB channels and calculate the PCA mean by sampling every third element along $h \times w$ vectors, each with feature dimension of either 512 (for LSeg) or 256 (for SAM). The feature map is transformed using the PCA components and mean. This involves centering the features with PCA mean and then projecting them onto PCA components. We then normalize the transformed feature based on the minimum and maximum values with the outliers removed to standardize the feature values into a consistent range so that it can be effectively visualized, typically as an image. We visualize both LSeg and SAM features of scenes from the LLFF dataset~\cite{mildenhall2019local} from different views. The feature map from LSeg encoder has size $360 \times 480$ with dimension $512$ (see in the second column of \cref{fig:feature_visualization}). However, as mentioned before, the feature map directly obtained from SAM encoder contains a padding region (see the red areas on the bottom of the feature maps in the third column), we crop out the region on the feature map as the ground-truth feature before feature distillation (see in the last column). It is worth noting that features from LSeg models mainly capture semantic information by delineating coarse-grained boundaries, while features from SAM models show instance-level information and even fine-grained details in different parts of an object. The capability of teacher encoders determines characteristics of the feature map, thereby influencing the upper limit of the performance of the rendered features on downstream tasks.

\section{Replica Dataset Experiment}
\label{sec:Replica Dataset Experiment}
Following the same selection in~\cite{stelzner2021decomposing}, we experiment on 4 scenes from the Replica dataset~\cite{straub2019replica}: room\_0, room\_1, office\_3, and office\_4. For each scene, 80 images are captured along a randomly chosen trajectory, and every 8th image starting from the third is selected. We trained 5,000 iterations on each scene with LSeg serving as the foundational model for this experiment. We manually re-label some pixels with semantically close labels such as ``rugs" and ``floor". This preprocess step follows the same method in the NeRF-DFF~\cite{kobayashi2022decomposing}. The model is trained on the training images and was subsequently evaluated on a set of 10 test images. We test pixel-wise mean intersection-over-union and accuracy on the manually relabeled test images and we use $class = 7$ for the mIoU metric. For room\_1, the last 2 test images are excluded from the results since these images do not have 7 classes in the image.
% \begin{table}[t]
% \begin{center}
% \fontsize{10pt}{12pt}\selectfont
% \resizebox{0.95\textwidth}{!}{
% \begin{tabular}{lccccccc} 
% \toprule
% Dimension & 8 & 16 & 32 & 64 & 128 & 256 & 512 \\
% \midrule
% Time (s)\downarrow & 6:40 & 7:21 & 8:51 & 12:10 & 19:55 & 48:39 & 1:29:42 \\
% mIoU\uparrow & 0.354 & 0.493 & 0.709 & 0.774 & 0.783 & 0.791 & 0.790 \\
% Accuracy\uparrow & 0.735 & 0.880 & 0.927 & 0.939 & 0.943 & 0.944 & 0.943 \\
% \bottomrule
% \end{tabular}}\vspace{-3mm}
% \caption{\textbf{Performance Metrics Across Different Dimensions.} The table shows the time, mean Intersection over Union (mIoU), and accuracy at various dimension levels.}
% \label{tab:performance_metrics}
% \end{center}
% \end{table}

% \begin{table*}[t]
% \begin{center}
% \fontsize{8pt}{7pt}\selectfont
% \resizebox{0.9\linewidth}{!}{
% \begin{tabular}{lccccccc} \toprule
% \toprule
% Dimension & 8 & 16 & 32 & 64 & 128 & 256 & 512 \\
% \midrule
% Time & 6:40 & 7:21 & 8:51 & 12:10 & 19:55 & 48:39 & 1:29:42 \\
% mIoU\uparrow & 0.354 & 0.493 & 0.709 & 0.774 & 0.783 & \textbf{0.791} & 0.790 \\
% Accuracy\uparrow & 0.735 & 0.880 & 0.927 & 0.939 & \textbf{0.944} & \textbf{0.944} & 0.943 \\
% \bottomrule
% \end{tabular}}
% \vspace{-3mm}
% \end{center}
% \caption{\textbf{Evaluation of Image Quality Metrics Across Different Dimensions.} This table presents the PSNR, SSIM, and LPIPS values corresponding to each dimension level.}
% \label{tab:performance_metrics}
% \end{table*}

\section{Editing Algorithm and Details}
\label{sec:Editing Algorithm and Details}

\begin{figure*}[t]
%   \fbox{\rule{0pt}{3in} \rule{0.9\linewidth}{0pt}}

  \includegraphics[width=\linewidth]{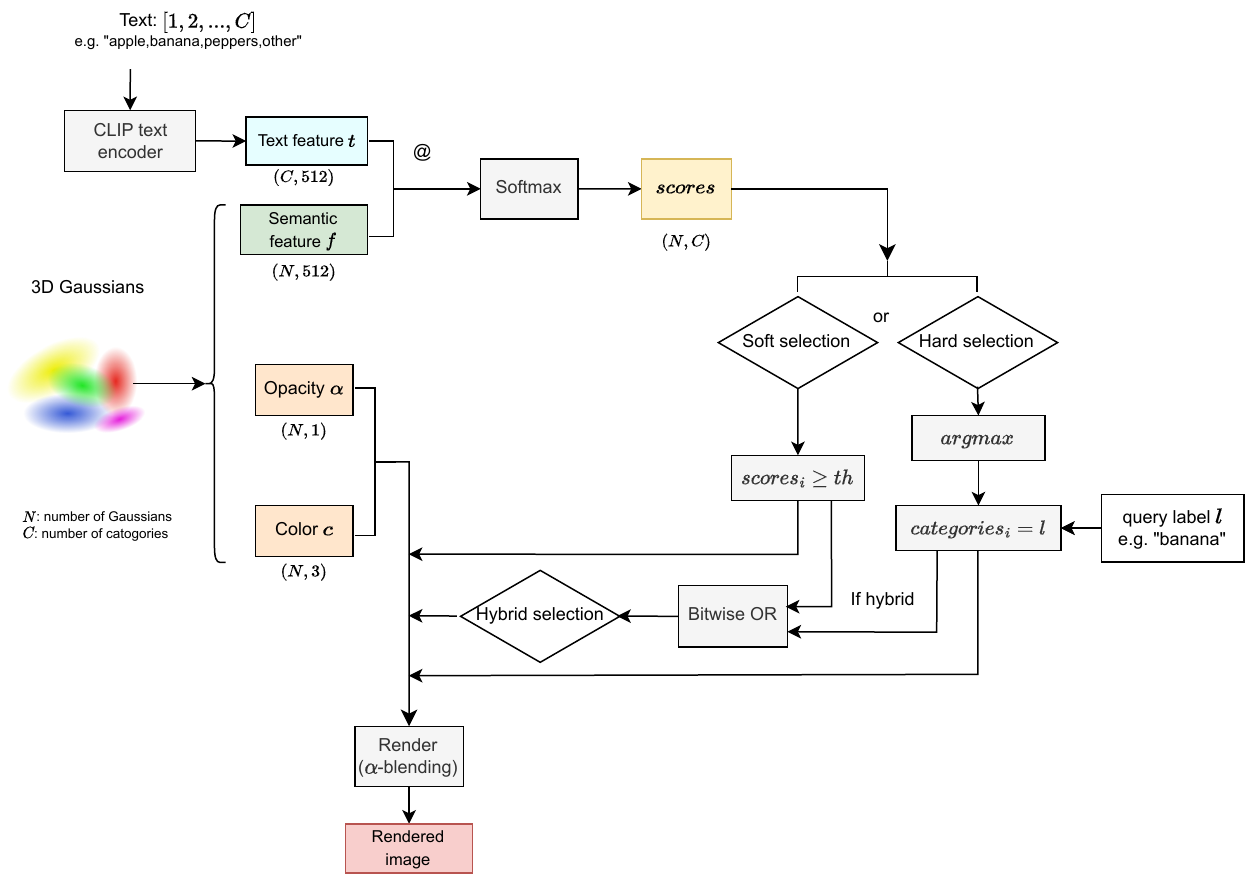}\vspace{-3mm}
  \caption{\textbf{Language-guided editing procedure using 3D Gaussians.} We calculate the inner product between the semantic feature and the text feature from CLIP encoder followed by a softmax to obtain a score matrix and query the feature field to apply editing on target regions (obtained from soft selection / hard selection / hybrid selection) by updating opacity and color from 3D Gaussians before rendering.} 
  \label{fig:edit_procedure}
\end{figure*}

The editing procedure takes advantage of the 3D Gaussians so that the model is able to render a novel view image edited with a specific editing operation. As illustrated in~\cref{fig:edit_procedure}, starting from a set of 3D Gaussians, i.e. $\mathcal{X} =\left\{x_1, \ldots, x_N\right\}$ where each $x_i$ is a 3D Gaussian represented by $(f_i, \alpha_i, c_i)$ where $f_i \in \mathbb{R}^{512}$, $\alpha_i \in \mathbb{R}$ and $c_i \in \mathbb{R}^{3}$ are the semantic feature, color and opacity, respectively. Guided by language, the edit algorithm takes a input text which is a list of object categories, e.g. `apple, banana, others'. We leverage the CLIP's ViT-B/32 text encoder for text encoding to obtain the text feature $\left\{t_1, \ldots, t_C\right\}$ where $t_i \in \mathbb{R}^{512}$ and $C$ is the number of categories. We then calculate the inner product of the text feature and semantic feature followed by a softmax function to obtain the semantic scores for each 3D Gaussian, represented by a $C$-dimensional vector, i.e. $scores \in \mathbb{R}^{N \times C}$. Queried by the text label $l$, e.g. `apple' or a list of objects to be edited, e.g. `apple, banana', one can either choose to apply hard selection or soft selection to perform edit operation specifically on the target region: 

\textbf{Soft selection:} Based on the category selected by the query label $l \in \left\{1, 2, \ldots, C\right\}$ (or $l \subseteq \left\{1, 2, \ldots, C\right\}$ if $l$ is a list of catogories), we target on the corresponding column of the score matrix, i.e. $score_l=\left[s_{1l}, s_{2l}, \ldots, s_{Nl }\right]^{\top}$ and apply binary thresholding on this column score vector, i.e. for any $i$ such that $s_{il} \geq th$, we set the position $i$ to 1 representing being selected; otherwise the position $i$ is set to 0 representing not being selected. Then all the positions $i$ such that $s_{li} = 1$ compose a target region to be edited. Intuitively, we mask out all the 3D Gaussians that are not selected and use those selected 3D Gaussians to update the color set \( \{c_i\}_{i=1}^N \) and opacity \( \{\alpha_i\}_{i=1}^N \).

\textbf{Hard selection:} We apply $argmax$ function to the score matrix to select the category corresponding to the highest score for each Gaussian to obtain a filtered category vector $categories = \left[c_1, c_2, \ldots, c_N\right]^{\top}$ where $c_i=\operatorname{argmax}\left\{s_{i 1}, s_{i 2}, \ldots, s_{i C}\right\}$. Then we filter based on the query label $l$: for any $i$ such that $c_{i} = l$ (or $c_i \in l$ if $l$ represents a list of target objects), we set the position $i$ to 1 representing being selected; otherwise the position $i$ is set to 0 representing not being selected. Similarly, we select the target region to be edited by preserving only the region positions of which the highest score category is aligned with the query label.

\textbf{Hybrid selection}: Since soft selection applies thresholding only based on column vector of score matrix corresponding to label $l$ which may potentially cause incorrect selection when the dominant score exists in other columns while hard selection merely selects the highest score without any tunable threshold value. Therefore, we propose a hybrid selection method by combining both hard and soft selection to alleviate the effect of incorrectly selecting the category while making the selection tunable to adapt to different scenarios. Spcifically, we combine the selection Gaussian masks of the two methods and apply bitwise OR operation between the two masks to obtain the final selected region.

We then update the opacity and color based on the selected edit region and a specific edit operation. We demonstrate the details of three examples: extraction, deletion and appearance modification:

\textbf{(a) Extraction:} for any $i \in \left\{1, \ldots, N\right\}$, if $i$ is selected, the opacity remains to be $\alpha_i$; otherwise the opacity is set to 0.

\textbf{(b) Deletion:} for any $i \in \left\{1, \ldots, N\right\}$, if $i$ is selected, the opacity is set to 0; otherwise the opacity remains to be $\alpha_i$. Specifically for deletion operation, we apply the hybrid selection method to select the target edit region to reduce the effect of incomplete deletion caused by the absence of target pixels.

\textbf{(c) Appearance modification:} for any $i \in \left\{1, \ldots, N\right\}$, if $i$ is selected, the color $c$ is updated to be $appearance \textunderscore func(c_i)$ where $appearance \textunderscore func(.)$ represents any appearance modification function, such as changing the green leaves to red leaves, etc.
%\clearpage

% \begin{figure}[t]
% %   \fbox{\rule{0pt}{3in} \rule{\linewidth}{0pt}}
%  \includegraphics[width=\linewidth]{figures/failure_cases.pdf}\vspace{-2mm}      
%      \vspace{-.2cm}\caption{\textbf{Failure results in complex and challenging situations} (a) The point-prompted segmentation mask, contains flaws in the form of small holes, resulting from low-quality features. (b) The model fails to delete tiny sophisticated objects thoroughly in a complex scene in language-guided editing.}
%   \label{fig:failure_cases}
% \end{figure}

\begin{figure*}[t]
  \centering
  \begin{minipage}{0.48\linewidth}
    \includegraphics[width=\linewidth]{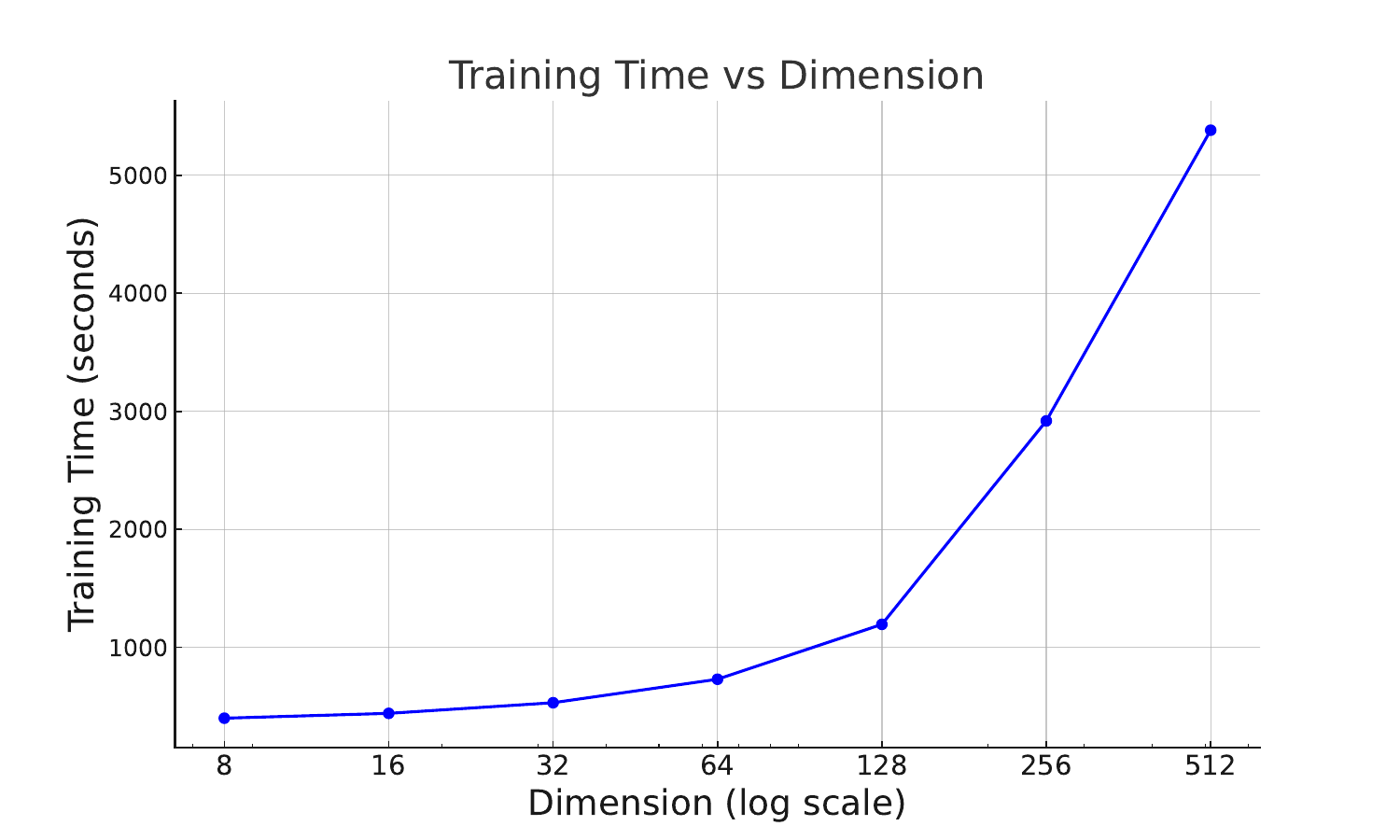}
    \caption{\textbf{Training Time vs Speed-up Module Dimension} We test the training time required with different input dimension of speed-up module. In this Figure, we show that the training time can be significantly reduced with our speed-up module.}
    \label{fig:Time_vs_Dimension}
  \end{minipage}\hfill
  \begin{minipage}{0.48\linewidth}
    \includegraphics[width=\linewidth]{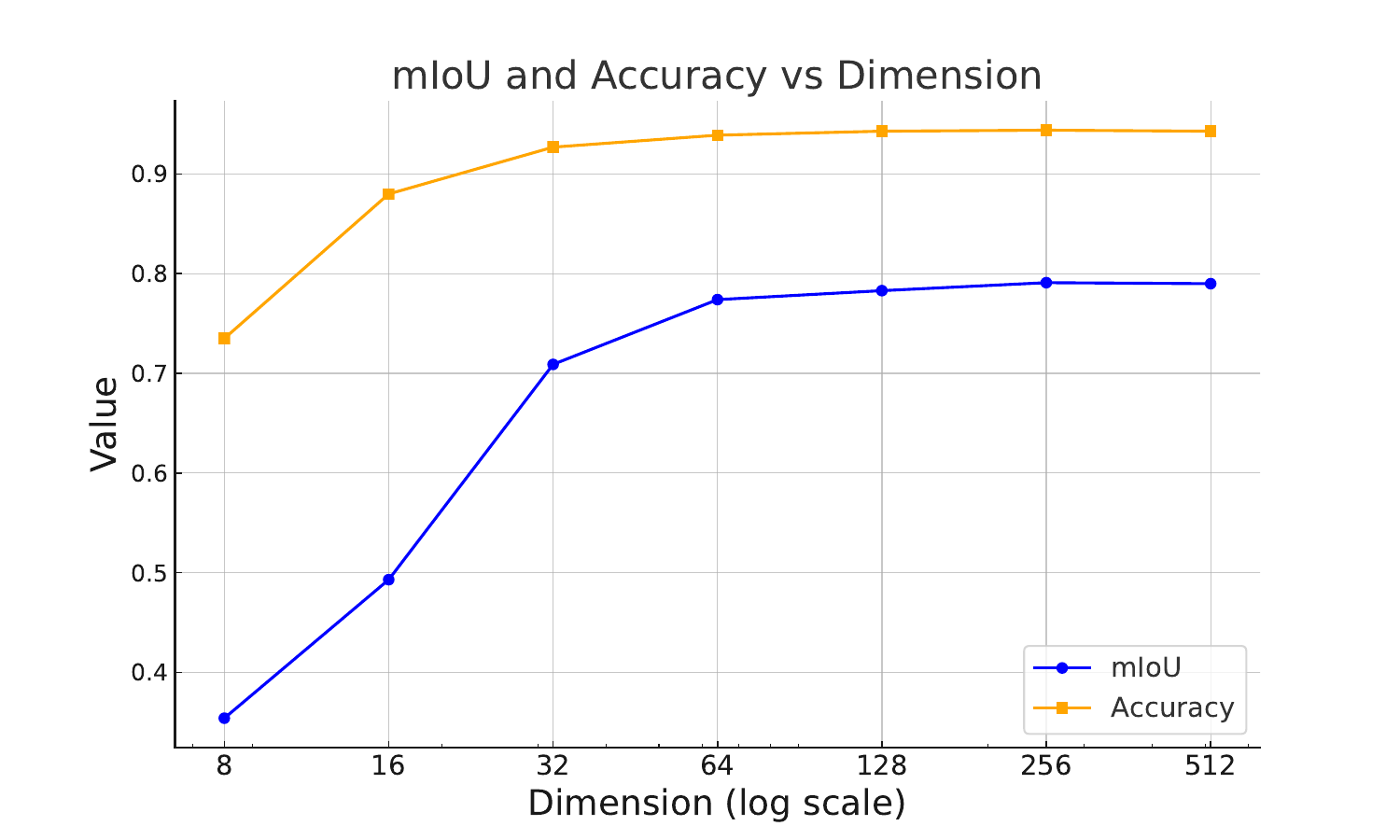}
    \caption{\textbf{mIoU and Accuracy vs Dimension} In this graph, we show 2D metrics with respect to different input dimensions of speed-up module. With our speed-up module and proper input dimension, the 2D metrics are not compromised.}
    \label{fig:mIoU_Accuracy_vs_Dimension}
  \end{minipage}
\end{figure*}

% \begin{figure}[t]
% %   \fbox{\rule{0pt}{3in} \rule{\linewidth}{0pt}}
%  \includegraphics[width=\linewidth]{figures/failure_cases.pdf}\vspace{-2mm}      
%      \vspace{-.2cm}\caption{\textbf{Failure results in complex and challenging situations} (a) The point-prompted segmentation mask, contains flaws in the form of small holes, resulting from low-quality features. (b) The model fails to delete tiny sophisticated objects thoroughly in a complex scene in language-guided editing.}
%   \label{fig:failure_cases}
% \end{figure}

\begin{figure*}[t]
%   \fbox{\rule{0pt}{3in} \rule{0.9\linewidth}{0pt}}

  \includegraphics[width=\linewidth]{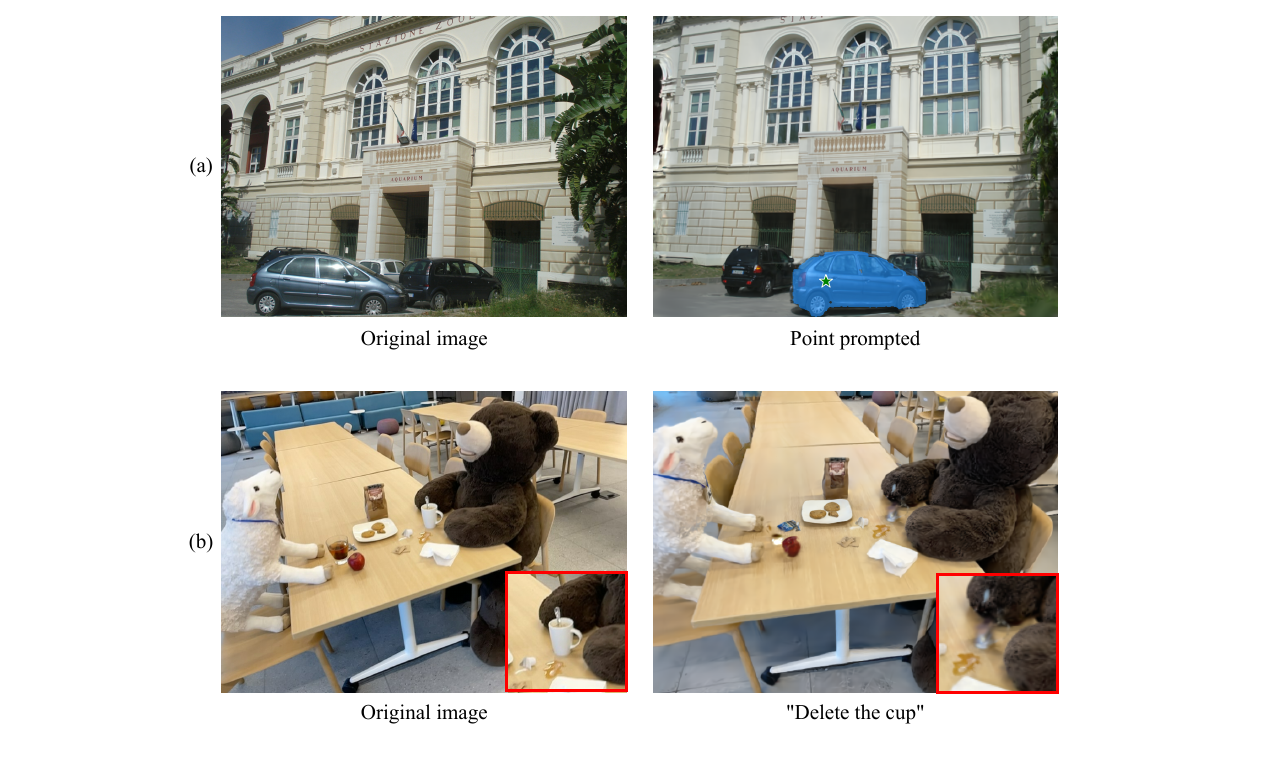}\vspace{-3mm}

  \caption{\textbf{Failure cases in complex and challenging situations} (a) The point-prompted segmentation mask, contains flaws in the form of a coarse boundary and small holes, resulting from low-quality features. (b) The model fails to delete tiny sophisticated objects thoroughly in a complex scene in language-guided editing.} 
  \label{fig:failure_cases}
\end{figure*}

\vspace{-3mm}
\section{Ablation Studies}
\vspace{-2mm}
\label{sec:Ablation Studies}

We study the effect of different rendered feature dimensions using our Speed-up Module. In~\cref{tab:performance_metrics}, we report the performance of the semantic segmentation on Replica dataset using LSeg feature. The result shows that both rendered feature $dim=256$ and $dim=128$ can achieve the best performance on accuracy, and $dim=256$ is slightly better on mIoU. However, $dim=128$ is $\times 2.4$ faster than $dim=256$ on training. We also report the quantitative results of novel view synthesis in~\cref{tab:image_quality_metrics_lseg}, which shows that $dim=128$ is the best. Therefore, we choose $dim=128$ for our Replica dataset experiment in practice. In addition, we show the performance and speed (FPS) of novel view synthesis with different dimensions of SAM feature in~\cref{tab:image_quality_metrics_sam}.

Furthermore,~\cref{fig:Time_vs_Dimension} and~\cref{fig:mIoU_Accuracy_vs_Dimension} substantiate that our Speed-up Module not only avoids compromising performance but, in fact, resulting in time savings.

\vspace{-3mm}
\section{Failure Cases}
\vspace{-2mm}
\label{sec:Failure Cases}
The proposed method indeed has limitations reflected on some failure cases. In~\cref{fig:failure_cases}, we showcase failure cases for scenes that are more challenging and complex. In~\cref{fig:failure_cases} (a), the point-prompted segmentation mask is not perfect with a coarse boundary and small holes. This is caused by low feature quality from SAM distillation, rather than Gaussian representation. Since the boundary of the car is hard to depicted and there are multiple similar objects close to each other (multiple adjacent cars), making the scene complex. As a result, achieving a smooth and accurate mask boundary of the car becomes challenging, which could be counted as a limitation. In~\cref{fig:failure_cases} (b), given the text prompt ``Delete the cup", although succeeding in locating the target object, the model fails to remove the cup comprehensively. The reason behind is that in some complex scenes including various objects with multiple sizes, the 3D Gaussians corresponding to the tiny objects with sophisticated details are hard to accurately selected by the ``Gaussian mask". As a result, a clean deletion is hard to perform on target object.

%\clearpage

% \clearpage
% %%%%%%%% REFERENCESsa
% {\small
% \bibliographystyle{ieee_fullname}
% \bibliography{supplement}
% }

% \clearpage
% {
% \small
% \bibliographystyle{ieeenat_fullname_sup}
% \bibliography{supplement}
% }

% \end{document}

\end{document}